\documentclass[10pt,twocolumn,letterpaper]{article}

\usepackage{sty_file/cvpr}
\usepackage{algorithm}
\usepackage{algorithmic}
\usepackage{amssymb,amsmath}
\usepackage{booktabs}
\usepackage[T1]{fontenc}
\usepackage{graphicx}
\usepackage{makecell}
\usepackage{multicol}
\usepackage{multirow}
\usepackage{pgfplots}
\usepackage{url}
\pgfplotsset{compat=1.18}

\definecolor{cvprblue}{rgb}{0.21,0.49,0.74}
\usepackage[pagebackref,breaklinks,colorlinks,allcolors=cvprblue]{hyperref}

\newtheorem{definition}{Definition}
\newtheorem{proposition}[definition]{Proposition}

\def\paperID{MUV-12} 
\def\confName{CVPR}
\def\confYear{2026}
\title{RUB: Evaluating Residual Knowledge in Unlearned Models}
\author{Hao Xuan \qquad Xingyu Li\\
Electrical and Computer Engineering\\
University of Alberta\\
{\tt \small \{hxuan,xingyu\}@ualberta.ca}
}
\begin{document}

\maketitle

\begin{abstract}

Machine Unlearning (MUL) has emerged as a key mechanism for privacy protection and content regulation, yet current techniques often fail to guarantee the complete removal of sensitive information. While most existing works focus on verifying the execution of unlearning, they overlook the critical question of whether models remain robust against adversarial attempts to recover forgotten knowledge. In this work, we advocate for the principle of Robust Unlearning, which requires models to be both indistinguishable from retrained counterparts and resilient against diverse adversarial threats. To instantiate this principle, we propose a unified benchmark, RUB (Robust Unlearning Benchmark), that systematically evaluates the robustness of unlearning algorithms across classification, image-to-image reconstruction, and text-to-image synthesis. Within this framework, we introduce the Unlearning Mapping Attack (UMA) as a generalizable method to detect residual information, and demonstrate how existing attack strategies can be adapted into this framework as long as they conform to the generic UMA framework. Our experiments across discriminative and generative tasks reveal that state-of-the-art unlearning methods remain vulnerable under these evaluations, even when passing standard verification metrics. By positioning robustness as the central criterion and providing a benchmark for adversarial evaluation, we hope RUB paves the way toward more reliable and secure unlearning practices. The codebase and model checkpoints in RUB will be published.

\end{abstract}

\section{Introduction}

As deep learning models grow increasingly data-dependent, concerns over privacy and data security have intensified. 
In response to privacy regulations such as the General Data Protection Regulation (GDPR)~\citep{GDPR} and the California Consumer Privacy Act (CCPA)~\citep{CCPA}, Machine Unlearning (MUL) has emerged as a potential solution to selectively remove specific data from trained models, allowing for the "right to be forgotten." Beyond privacy concerns, content regulation has become another key motivation for machine unlearning~\citep{kurmanji2024towards,shumailov2024ununlearning}. To remove impermissible knowledge, such as unlicensed copyrighted material~\citep{yao2023large}, malicious information~\citep{yao2023large}, or harmful capabilities~\citep{shumailov2024ununlearning} from models, machine unlearning ensures that the unlearned models align with ethical and legal standards. To date, MUL techniques have demonstrated strong performance in eliminating the influence of specific data on both privacy-sensitive and content-sensitive tasks~\citep{warnecke2021machine,li2024machine,graves2021amnesiac, tarun2023fast, Golatkar_2020_CVPR,liu2022right}. 

Since MUL handles sensitive data, it is inherently vulnerable to adversarial attacks. Prior research has shown that modifying the unlearning process or manipulating training data, such as injecting noise or backdoors, can undermine its effectiveness~\cite{thudi2022necessity,qian2023towards,liu2024backdoor,zhangverification}. However, beyond these conventional attacks, a new category of post-unlearning adversarial attacks has emerged, targeting residual information left in unlearned models~\citep{zhang2025generate,tsai2023ring,han2024probing,pham2023circumventing,yuan2024towards}. Though these attacks are particularly designed for diffusion models (DMs) or large language models (LLMs) to fail content erasure, they expose a fundamental vulnerability: unlearned models often retain traces of the forgotten data, which adversarial probes can exploit to resurface unlearned information. In other words, attackers can craft adversarial inputs that recover forgotten knowledge, effectively negating the unlearning process.

To evaluate MUL effectiveness and robustness, verification methods typically fall into attack-based empirical evaluation and process-based reproducibility checks. The former tests an unlearned model’s resistance to adversarial threats by either extracting forgotten information~\citep{fredrikson2015model,shokri2017membership,carlini2022membership} or injecting backdoors to deceive the model~\citep{sommer2020towards, gao2024verifi, 10298847}. The latter, inspired by Proof of Learning (PoL)~\citep{jia2021proof}, logs the unlearning process, allowing users or auditors to verify whether unlearning was executed~\citep{zhangverification}. Despite their usefulness, these methods focus primarily on whether unlearning was performed, rather than ensuring that all traces of the unlearned information have been irreversibly removed. 

To address this critical gap in MUL, we introduce the first framework, namely Robust Unlearning Benchmark (RUB), that systematically evaluates the robustness of machine unlearning against post-unlearning adversarial information resurface over various computer vision tasks. As demonstrated in Fig.~\ref{fig:threat_model}, RUB covers 15 unlearning algorithms, 3 computer vision tasks (i.e. image classification, image-to-image reconstruction, and text-to-image synthesis) over diverse benchmarking datasets, and different post-unlearning adversarial probing methods. At its core, RUB defines a unified protocol for evaluating whether forgotten information can be recovered under attacks. Particularly, we propose a generic Unlearning Mapping Attack (UMA) as a modular adversarial probing tool, and instantiate it across three major task settings. 
Our benchmark includes multiple evaluation metrics tailored to different tasks and reveals striking differences in the robustness of existing unlearning methods. The benchmarking results on the 15 unlearning methods and our in-depth discussions highlight potential research directions in the future. Our contributions are summarized as follows:

\begin{figure*}[t!]
    \centering
    \includegraphics[width=\linewidth]{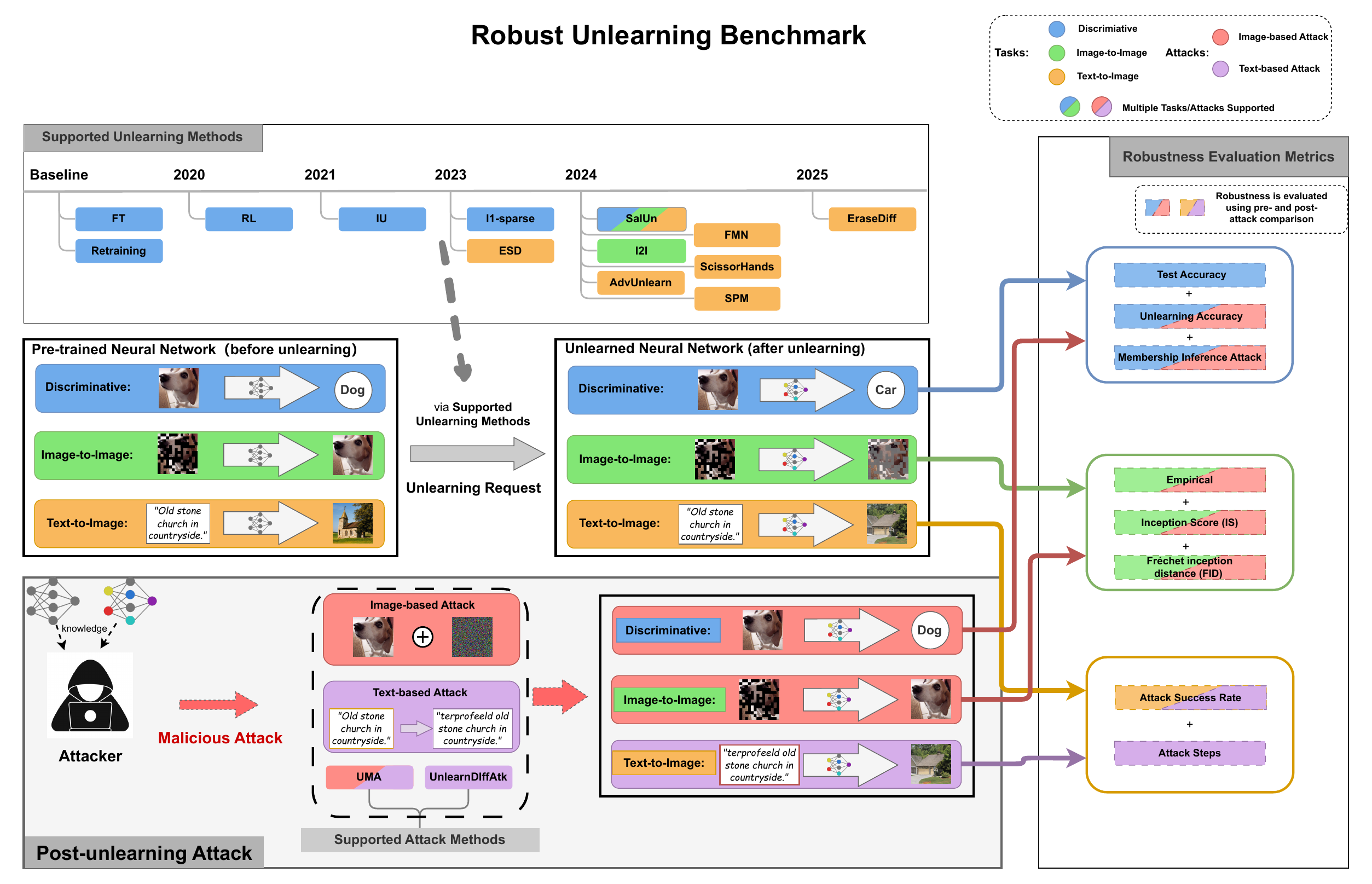}
    \caption{Diagram of our robust unlearning benchmark. RUB covers three common unlearning tasks: classification, image-to-image reconstruction, and text-to-image synthesis, each reflecting distinct unlearning objectives. It supports multiple state-of-the-art unlearning algorithms, task-specific white-box adversarial attack implementations, and appropriate evaluation metrics. Our benchmark provides a unified and extensible framework to assess the robustness of unlearning algorithms against malicious recovery of unlearned information.}
    \label{fig:threat_model}
\end{figure*}

\begin{itemize}
    \item We introduce the first benchmark to systematically evaluate unlearning algorithms under post-unlearning adversarial recovery attacks across diverse computer vision tasks.
    \item We have created a unified evaluation protocol and well-structured code-base that supports 15 state-of-the-art unlearning algorithms on post-unlearning adversarial robustness.
    \item We design UMA, a general-purpose adversarial attack that can be instantiated across tasks, providing an empirical verification tool to assess whether unlearning methods completely eliminate residual knowledge.
    \item We have conducted a thorough analysis of various unlearning algorithms in our benchmark, which will inspire researchers to develop more robust unlearning algorithms.
\end{itemize}

\section{Related Work}
\textbf{Machine Unlearning.} Machine Unlearning~\citep{Cao2015} was introduced to remove specific data influences from models for privacy and security. The most straightforward approach, retraining from scratch, has become impractical for large models like LLMs due to their high computational cost. To address this, exact and approximate unlearning are proposed. Exact unlearning~\citep{Golatkar_2020_CVPR,bourtoule2021machine} aims to make an unlearned model indistinguishable from one retrained from scratch without the forgotten data. 
Given scalability challenges, approximate unlearning offers a more practical alternative by relaxing constraints to improve efficiency while maintaining acceptable performance. First-order methods use Taylor series expansions to update model parameters, while second-order methods incorporate the inverse Hessian matrix for more precise adjustments~\citep{warnecke2021machine}. Recently, SalUn~\citep{fan2024salun} improved stability and accuracy by using a weight saliency map to update parameters at varying rates. Despite these advancements, existing methods remain vulnerable to sophisticated attacks that extract forgotten information, raising critical privacy and security concerns.



\textbf{Attacks to MUL.} Machine unlearning is inherently vulnerable to adversarial attacks, as evidenced by prior research on malicious attempts against MUL. For instance, adversarial text prompts and Concept Inversion attacks are particularly investigated to undermine DMs for content erasure~\citep{zhang2025generate,tsai2023ring,han2024probing,pham2023circumventing}. Targeting more generic unlearning scenarios, \cite{qian2023towards} injects targeted noise into forget samples, leading the unlearned model to fail in classification tasks. 
\cite{liu2024backdoor} induces backdoor behavior in a model through the standard MUL process with selected data. 
\cite{thudi2022necessity,zhangverification} 
leverage techniques from Data Ordering Attacks\citep{shumailov2021manipulating} to falsify Proof-of-Unlearning (PoUL), with the intent of either enhancing model performance or reducing computational costs. 

\section{Robust Machine Unlearning: Preliminaries}



\textbf{Notation.} Let $\mathcal{D}=\{(x_i,y_i)^N_{i=1}\}$ denote the training dataset used to train a model $f(\cdot; \theta)$, where $N$ represents the number of training samples, each consisting of input features $x \in \mathbb{R}^d$ and target output $y$, and $\theta$ denotes the model's parameters. The model's training process is represented as $\mathcal{A}(f(\cdot; \theta), \mathcal{D})$, while the unlearning process is denoted by $\mathcal{U}(f(\cdot; \theta), \mathcal{D}_u)$, where $\mathcal{D}_u \subset \mathcal{D}$ is the subset of data to be unlearned. After the unlearning process, the updated model is expressed as $f_u(\cdot; \theta^u)$, i.e., $f_u(\cdot; \theta^u) = \mathcal{U}(f(\cdot; \theta), \mathcal{D}_u)$.

\begin{definition}\label{def_ul}
    ({Machine Unlearning}~\citep{Cao2015}.)
    An unlearning process $\mathcal{U}(f(\cdot;\theta),\mathcal{D}_u)$ aims to find an unlearned model $f_u(\cdot;\theta^u)$ so that it closely aligns with a model trained from scratch on the retain set $\mathcal{D}_r=\mathcal{D}/\mathcal{D}_u$, i.e. $f_u(\cdot;\theta^u)=f_r(\cdot;\theta^r)=\mathcal{A}(f(\cdot;\theta),\mathcal{D}_r)$.
\end{definition}


According to Definition \ref{def_ul}, for MUL evaluation, the performance of an unlearning algorithm should closely align with that of a retrained model on the retain set. In practice, especially for large-scale systems like foundation models, where retraining is computationally prohibitive or impractical for information unlearning, a more feasible approach is to ensure its performance has sufficient divergence from the original model on the forget set while maintaining performance on the retain set. For example, a generative unlearned model should no longer be capable of producing undesirable information from the forget set~\citep{warnecke2021machine,fan2024salun,li2024machine}. That is, {\color{black}{empirically, an unlearned model should decorrelate the input $x \in \mathcal{D}_u$ from the original output by a significant margin $\varepsilon_1$, 
while maintaining close predictions for $x \in \mathcal{D}_r$ for its utility:
\begin{eqnarray}
\left\{
    \begin{array}{cc}
E_{x\in\mathcal{D}_u}[||f_u(x,\theta^u)-f(x,\theta)||]>\varepsilon_1 \\
E_{x\in\mathcal{D}_r}[||f_u(x,\theta^u)-f(x,\theta)||]\leq\varepsilon_2
    \end{array}
    \right.
\label{eq_unlearn_dist_2}
\end{eqnarray}
Here $E$ represents the statistical mean over a distribution, and $\varepsilon_1$ and $\varepsilon_2$ may vary depending on the task and should align with real-world attack detectability and interpretability of human visual inspection. These empirical observations in~(\ref{eq_unlearn_dist_2}) underpin the commonly used performance metrics for evaluating unlearning, typically focusing on performance over the retrain set and the forgetting set. However, such evaluations overlook a critical security gap - the possibility of adversarial recovery of forgotten information by arbitrary or adversarial crafted perturbations~\citep{zhang2025generate,tsai2023ring,han2024probing,pham2023circumventing,yuan2024towards} - and therefore fail to comprehensively assess the true effectiveness of unlearning models. 
In this study, we formulate this vulnerability of MUL in knowledge regulation and information removal as the following two propositions:
\begin{proposition}\label{prop_risk}
    {\color{black}{For an unlearned generative system $f_u(\cdot,\theta^u)$ that satisfies conditions in (\ref{eq_unlearn_dist_2}), there may exist an adversarial probing $\delta_x \notin \mathcal{D}_u$ s.t. $||f_u(\delta_x,\theta^u)-f(x,\theta)||<\varepsilon_1, \forall x\in \mathcal{D}_u$.}}
\end{proposition}
\begin{proposition}\label{prop_risk_D}
    {\color{black}{For an unlearned discriminative model $f_u(\cdot,\theta^u)$ that satisfies conditions in (\ref{eq_unlearn_dist_2}), there may exist a small non-zero $\delta_x \in \mathbb{R}^d$ (i.e. $||\delta_x|| < \epsilon$)  s.t. $||f_u(x+\delta_x,\theta^u)-f(x,\theta)||<\varepsilon_1, \forall x\in \mathcal{D}_u$.}}
\end{proposition}
Proposition~\ref{prop_risk} considers scenarios where a generative model produces outputs based on various prompts or inputs, and Proposition~\ref{prop_risk_D} focuses on discriminative tasks where slight perturbations $\delta_x$ on the data $x$ can bypass the unlearning process and the system’s security is compromised. Note, $\delta_x$ in both Propositions is data-specific and model-specific. The extra constraint $||\delta_x|| < \epsilon$ for discriminative models ensures semantic similarity between $x$ and $x + \delta_x$, so that the attack remains meaningful and realistic, aligning with its practical use of these models. By contrast, generative models allow for unconstrained $\delta_x$, as the attack focuses exclusively on the outputs generated from crafted inputs, irrespective of input realism.

Since machine unlearning is inherently tied to data security, unlearning algorithms must be robust and resilient to such malicious recovery attempts. 


\begin{proposition}\label{def_rob}
(Robust Unlearning). A unlearning process, $\mathcal{U}(f(\cdot;\theta),\mathcal{D}_u)$, is considered robust if $\forall x\in \mathcal{D}_u$, $\forall \delta_x \in \mathbb{R}^d$, we have conditions in (\ref{eq_unlearn_dist_2}), plus $||f_u(\delta_x,\theta^u)-f(x,\theta)|| > \varepsilon_1$ for generative tasks and $||f_u(x+\delta_x,\theta^u)-f(x,\theta)|| > \varepsilon_1$ for discriminative models (where $||\delta_x||<\epsilon$).
\end{proposition}

Intuitively, Robust Unlearning ensures that the system is incapable of producing the specified information, whether under normal conditions or in the presence of adversarial manipulation. We propose this as a comprehensive and robust standard for defining and evaluating unlearning algorithms in our benchmark. 
For instance, our empirical experiments on discriminative models in Table \ref{tab:class_unlearning} show that retraining achieves strong robustness, while those evaluated unlearning algorithms are less robust to adversarial attacks, suggesting these vulnerabilities are intrinsic to the unlearning methods. 

\section{RUB Benchmarking Framework}

RUB aims to systematically evaluate the robustness of machine unlearning algorithms against adversarial recovery of forgotten information. To this end, we introduce a unified evaluation protocol, which probes whether unlearned models still retain residual information of the unlearned data. To ensure broad applicability, our benchmark spans three distinct tasks: classification, image-to-image reconstruction, and text-to-image synthesis, each representing a different modality and unlearning objective. For each task, we define tailored adversarial attack strategies and evaluation metrics to quantify the model's susceptibility to post-unlearning information recovery. Note, to evaluate the upper bound of recoverable information from unlearned models, we adopt a white-box, worst-case assumption for the adversary attack in our benchmark, that is, RUB has full access to the original model \textit{before} and \textit{after} the unlearning process, along with knowledge of the specific samples designed for removal.



\subsection{Unlearning Scenarios/Tasks}
To comprehensively evaluate the robustness of unlearning algorithms, RUB spans multiple unlearning scenarios. We categorize these into three representative tasks highlighted by blue, green, and orange in Fig.~\ref{fig:threat_model}, respectively. Each of these tasks targets a fundamentally different form of information retention and thus presents unique challenges for unlearning.

In \textbf{discriminative unlearning}, unlearning typically involves removing the influence of specific training samples or entire classes from a discriminative model. This setting is well-studied in terms of classification and forms the foundation of many early unlearning algorithms. 
Our benchmark includes classification unlearning as a canonical use case and evaluates both natural forgetting efficacy and adversarial vulnerability. Specifically, following prior arts~\citep{fan2024salun}, our classification unlearning evaluation will be performed on 3 datasets: \textit{CIFAR10, CIFAR100, and Tiny-ImageNet}.

\textbf{Image-to-Image (I2I) reconstruction unlearning} targets to erase a model’s ability to reconstruct specific visual content, while preserving its performance on the retained dataset. To instantiate this task, we follow the prior unlearning study SalUn~\citep{fan2024salun} for masked image inpainting, where a generative model is pre-trained to reconstruct missing regions from partially masked inputs. After unlearning the forget data, the model's behavior is then assessed by feeding in masked versions of those forgotten images. If unlearning is successful, the model should fail to accurately reconstruct the masked regions, instead producing uninformative or generic outputs. Any successful recovery of the original content, particularly under adversarial probing, suggests residual memorization and a failure to fully forget. In the I2I scenario, we adopt \textit{ImageNet-1k} as the benchmarking set.

\textbf{Text-to-Image (T2I) synthesis unlearning} aims to remove the generative model’s ability to produce images corresponding to textual prompts. With the rise of powerful DMs, such safety-driven unlearning has become increasingly important. Usually, the forgetting targets are abstract and semantic, such as high-level concepts embedded in a diffusion model. In this task, RUB leverages the unlearning codebase in UnlearnDiffAtk~\citep{zhang2024generate} for this purpose. Specifically, unlearning algorithms are applied to the pre-trained DM for the provided prompt lists in UnlearnDiffAtk, and we evaluate whether unlearned DMs can still be coerced into generating "forgotten" content. 

\begin{table*}[t]
    \centering
    \small
    \setlength{\tabcolsep}{12pt}
    \begin{tabular}{cccccccc}
        \toprule
        \multirow{2}[2]{*}{\textbf{CIFAR10}}
        & \multicolumn{3}{c}{No Atk}& \multicolumn{2}{c}{$\epsilon=8/255$}& \multicolumn{2}{c}{$\epsilon=16/255$}\\
        \cmidrule(lr){2-4} \cmidrule(lr){5-6} \cmidrule(lr){7-8}
        & TA $\uparrow$& UA $\downarrow$& MIA$\downarrow$& UA$\downarrow$& MIA$\downarrow$& UA$\downarrow$& MIA$\downarrow$\\
        \hline\toprule
        Original& 94.13& 100& 0.9796& -& -& -& -\\
        retrain& 94.14& 0& 0& 0& 0& 0& 0\\
        FT& 91.82& 20.55& 0.088& 99.96& 0.995& 99.98& 0.995\\
        RL& 92.19& 0& 0& 5.82& 0.024& 26.64& 0.135\\
        IU& 88.06& 8.76& 0.058& 99.12& 0.965& 99.87& 0.983\\
        $l_1$-sparse& 90.00& 0& 0& 98.30& 0.871& 99.90& 0.978\\
        SalUn& 92.70& 0& 0& 7.13& 0.036& 26.87& 0.148\\
        \toprule
        \multirow{2}[2]{*}{\textbf{CIFAR100}}
        & \multicolumn{3}{c}{No Atk}& \multicolumn{2}{c}{$\epsilon=8/255$}& \multicolumn{2}{c}{$\epsilon=16/255$}\\
        \cmidrule(lr){2-4} \cmidrule(lr){5-6} \cmidrule(lr){7-8}
        & TA$\uparrow$& UA$\downarrow$& MIA$\downarrow$& UA$\downarrow$& MIA$\downarrow$& UA$\downarrow$& MIA$\downarrow$\\
        \hline\toprule
        Original& 75.25& 100& 0.9908& -& -& -& -\\
        retrain& 75.40& 0& 0.024& 0& 0.014& 0& 0.014\\
        FT& 67.64& 0.48& 0.306& 99.28& 0.977& 99.89& 0.992\\
        RL& 69.96& 3.20& 0.269& 51.53& 0.688& 80.50& 0.790\\
        IU& 66.42& 53.37& 0.848& 99.93& 1& 99.93& 1\\
        $l_1$-sparse& 70.70& 1.30& 0.402& 99.77& 0.925& 99.91& 0.945\\
        SalUn& 73.89& 4.13& 0.221& 61.57& 0.788& 85.16& 0.888\\
        \toprule
        \multirow{2}[2]{*}{\textbf{Tiny-ImageNet}}
        & \multicolumn{3}{c}{No Atk}& \multicolumn{2}{c}{$\epsilon=8/255$}& \multicolumn{2}{c}{$\epsilon=16/255$}\\
        \cmidrule(lr){2-4} \cmidrule(lr){5-6} \cmidrule(lr){7-8}
        & TA$\uparrow$& UA$\downarrow$& MIA$\downarrow$& UA$\downarrow$& MIA$\downarrow$& UA$\downarrow$& MIA$\downarrow$\\
        \hline\toprule
        Original& 64.17& 99.96& 1& -& -& -& -\\
        retrain& 57.74& 0& 0& 0& 0& 0& 0\\
        FT& 60.48& 79.01& 0.721& 99.99& 0.991& 99.99& 0.991\\
        RL& 56.23& 2.09& 0.028& 99.78& 0.859& 99.99& 0.917\\
        IU& 57.71& 94.44& 0.882& 99.99& 0.991& 99.99& 0.991\\
        $l_1$-sparse& 58.28& 45.99& 0.228& 99.99& 0.833& 99.99& 0.843\\
        SalUn& 57.82& 5.95& 0.084& 99.98& 0.964& 99.99& 0.982\\
        \bottomrule
    \end{tabular}
    \caption{Test Accuracy (TA), Unlearning Accuracy (UA), and MIA scores before and after adversarial attacks for the Class Unlearning scenario. Attack is bounded with 8/255 and 16/255. The original here indicates the model performance before unlearning.}
    \label{tab:class_unlearning}
\end{table*}

\subsection{Evaluation Protocol and Metrics}

\textbf{Evaluation protocol.} In our benchmark, a strong adversarial evaluation protocol is designed to assess the robustness of unlearning algorithms against malicious recovery of forgotten information. 
To evaluate the upper bound of recoverable information from unlearned models, we adopt a white-box, worst-case assumption for the adversary, that is, RUB has full access to the original model \textit{before} and \textit{after} the unlearning process, along with knowledge of the specific samples designed for removal. As shown in Fig.~\ref{fig:threat_model}, for a given unlearning task, we begin with a pretrained model trained on the full dataset (i.e. $D_u\cup D_r$). An unlearning algorithm is then applied to remove the influence of target data, resulting in an unlearned model. To assess adversarial robustness, we craft a probing attack (details in Section~\ref{UMAttack}) to actively resurface forgotten information. We then evaluate the post-attack performance on the forgetting set.  

\textbf{Evaluation metrics.} 
Since unlearning objectives vary significantly across tasks, there are no universally applicable metrics. As such, we design task-specific evaluation metrics as follows. Please refer to the Appendix for detailed information.

For classification unlearning, we utilize both \textit{Unlearning Accuracy} (UA) and \textit{Membership Inference Attack} (MIA) as the evaluation metrics. It is important to clarify the interpretation of UA. In some prior works, UA has been regarded as the unlearning success rate, where higher values indicate better performance. In contrast, we define UA as the model’s raw accuracy on the forget dataset, which should be interpreted as lower values indicating better unlearning. For the MIA evaluation, we adopt a shadow-model-based MIA strategy~\cite{shokri2017membership} (See details in the Appendix). To address the randomness in MIA for reliable evaluation, we randomly sampled 10 fixed random seeds for executing unlearning and 5 fixed random seeds for training MIA attack models. In total, we have 50 sets of results, and their statistics are reported. 

For I2I unlearning, we perform both qualitative and quantitative evaluations. Qualitatively, human perception plays a key role, and the primary question is whether the generated image appears visually similar to the original. Quantitatively, we adopt standard image generation metrics, including the Inception Score (IS)~\citep{salimans2016improved} and the Fréchet Inception Distance (FID)~\citep{heusel2017gans}, to measure output quality. A higher IS and a lower FID indicate better generation performance.

For T2I unlearning, we evaluate robustness using two key indicators: the attack success rate and the number of steps required to achieve a successful attack. These metrics reflect how much forgotten information remains recoverable, providing a quantitative measure of the unlearning model's resilience under adversarial probing. To determine attack success, we follow the protocol from UnlearnDiffAtk~\citep{zhang2024generate}, employing a classifier to assess whether the generated image is recognized as belonging to the forgotten class.

\subsection{Supported Unlearning Algorithms}
Our benchmark contains 15 unlearning algorithms, ranging from MUL baseline approaches to the latest DM-specific unlearning methods, among which 6 are applicable to classification unlearning, 2 to I2I unlearning, and 7 to T2I unlearning. Specifically, our classification unlearning can be achieved by  FT~\citep{warnecke2021machine}, RL~\citep{Golatkar_2020_CVPR}, IU~\citep{koh2017understanding,izzo2021approximate}, l1-sparse~\citep{jia2023model}, and SalUn~\citep{fan2024salun}. The two unlearning methods applicable to I2I unlearning in literature are I2I~\citep{li2024machine} and SalUn~\citep{fan2024salun}. For T2I unlearning, various DM-tailored algorithms, AdvUnlearn~\citep{zhang2024defensive}, EraseDiff~\citep{wu2025erasing}, ESD~\citep{gandikota2023erasing}, FMN~\citep{zhang2024forget}, SalUn~\citep{fan2024salun}, ScissorHands~\citep{wu2024scissorhands}, and SPM~\citep{lyu2024one}, are evaluated in this benchmark.

\subsection{Adversarial Recovery Attacks}\label{UMAttack}

One key module in RUB is the adversarial attack that is able to exploit
residual knowledge within the unlearned model for forgotten information resurfacing. This process reveals the extent to which existing unlearning techniques eliminate traces of forgotten data, providing a direct empirical measure of unlearning robustness. Although several adversarial recovery attacks have been proposed for DMs and LLMs, they are not applicable to the canonical discriminative unlearning and I2I unlearning scenarios. To this end, we introduce a generic white-box post-unlearning attack prototype, namely Unlearning Mapping Attack. It should be noted that though UMA is introduced as the default attack module in this benchmark, other attacks can be easily plugged in.
\begin{definition}\label{def_UMA}
    (Generic UMA) In the context of MUL, an adversarial strategy is considered UMA given that the attack, by \textbf{modifying the input} to a model, causes the output to \textbf{reveal or approximate} information or concept that was intended to be removed or forgotten through the unlearning process.
\end{definition}
Notably, UMA probes an unlearning model by varying its input in inference only; It does not require any change to the unlearning algorithm or the model parameters after the unlearning process. Let $D$ quantifies the behavior difference of two models, a white-box UMA can be formulated as
\begin{equation}
\mathop{\arg\min}_{\delta_i}D[f_u(\delta_i,\theta^u),f(x_i,\theta)], \forall x_i\in \mathcal{D}_u.
\label{eq_goal}
\end{equation}
Here, $D$ can vary depending on the context, such as the KL Divergence on classification logits, and the Mean Square Error for I2I reconstruction. The UMA formulation in Eq.~(\ref{eq_goal}) aligns with the proposition of robust unlearning. If for every $x\in \mathcal{D}_u$ we find an optimal $\delta_x$ to minimize the difference, and the minimum difference is still larger than $\varepsilon_1$, we can conclude that the unlearned model is robust with respect to $\varepsilon_1$. 

Unlike the models in the first two unlearning tasks, which produce outputs in a single forward pass, DMs in T2I unlearning are inherently iterative generative processes. Accordingly, rather than comparing pre-attack and post-attack image outputs for UMA calculation, we can leverage DM's training mechanism, i.e. noise level estimation, as the quantitative performance measurement in Eq.~(\ref{eq_goal}). Under this formulation, the UMA attack for T2I tasks is instantiated as follows.
\begin{equation}
\mathop{\arg\min}_{\delta_i}E_{t}||\epsilon_{\theta^u}(x_i^t|\delta_i)-\epsilon_{\theta}(x_i^t|c)||, \forall x_i\in \mathcal{D}_u,
\label{eq_DMUMA}
\end{equation}
where $E_t$ indicates that the attack is optimized over all $t$ steps in the diffusion process, and $c$ and $\delta_i$ are the original and adversarial textual prompts, respectively. It is worth noting that the UMA objective in Eq.~(\ref{eq_DMUMA}) closely resembles that of existing adversarial attacks on unlearned diffusion models~\citep{tsai2023ring, zhang2024generate}. Therefore, we adopt the gradient-based attack in UnlearnDiffAtk~\citep{zhang2024generate} in our benchmark to recover forgotten information.


\section{Experiments and Discussions}
\subsection{Implementation Details}
\textbf{Discriminative unlearning.} We choose ResNet50 as our model backbone and conduct class-wise unlearning experiments on all three datasets. For each dataset, we randomly pick 10\% of the total classes to evaluate unlearning performance. In our case, class 0 is selected for CIFAR10, class 1, 2, 24, 27, 41, 50, 52, 73, 78, 91 for CIFAR100, and class 3, 11, 17, 34, 49, 59, 97, 107, 109, 129, 133, 137, 154, 156, 173, 177, 179, 183, 194, 197 for Tiny-ImageNet as the unlearning target. Since the attack for discrimination tasks unlearning needs to be bounded, or it would be meaningless otherwise, we bound the perturbation to 8/255 and 16/255, which represent the maximum noise strength while preserving major feature information. 


\textbf{Image-to-Image unlearning.} We follow the study in I2I attack~\citep{li2024machine} and adopt class unlearning using Masked AutoEncoder (MAE) on ImageNet-1k dataset. We utilize the class index file from I2I attack project page as our dataset, which contains 200 class indices from ImageNet-1k. The first 100 classes are selected to be the forget set, while the last 100 classes are the retain set.

\textbf{Text-to-Image unlearning.} We evaluate a range of T2I unlearning methods on the task of forgetting specific objects and concepts. Following prior work in T2I unlearning~\citep{zhang2024generate}, RUB includes the following target concepts for removal: Church, Garbage Truck, Nudity, Parachute, and Tench. To compute the average number of steps required to compromise unlearning, we cap the maximum number of adversarial prompt optimization iterations at 100.

We also present full details of our implementation, such as unlearning hyperparameters and checkpoints we used for evaluation in the Appendix.

\begin{figure*}[t]
    \centering
    \includegraphics[width=0.98\linewidth]{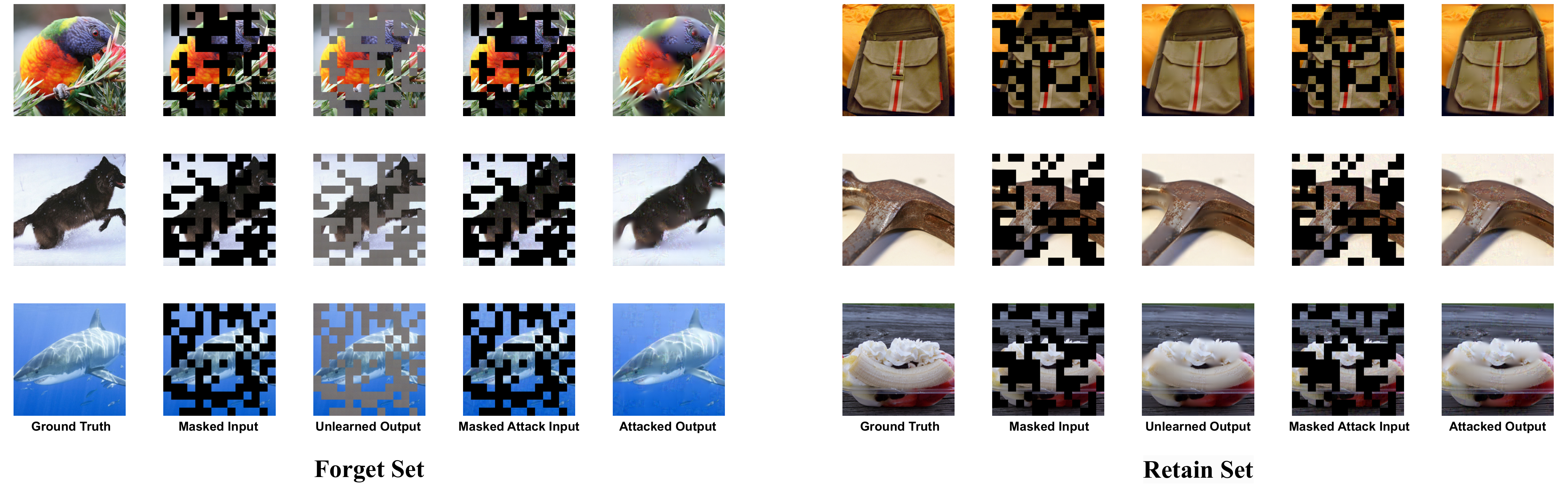}
    \caption{Unlearning Mapping Attack on image generation unlearning. I2I~\citep{li2024machine} unlearning method is tested here. Reconstructed images are from ImageNet1k dataset.}
    \label{fig:mae_result1}
\end{figure*}
\begin{figure*}[htbp]
    \centering
    \includegraphics[width=0.98\linewidth]{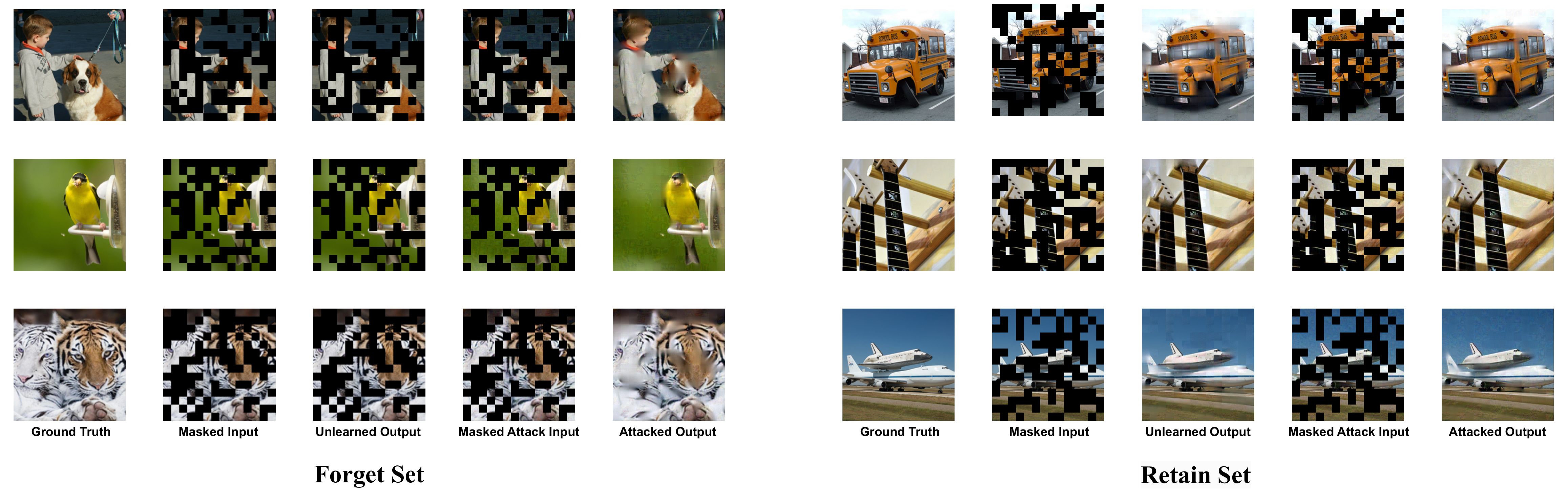}
    \caption{Unlearning Mapping Attack on image generation unlearning. SalUn~\citep{fan2024salun} unlearning method is tested here. Reconstructed images are from ImageNet1k dataset.}
    \label{fig:mae_result2}
\end{figure*}

\subsection{Evaluation Results and Discussions}
\textbf{Discriminative unlearning.} Table \ref{tab:class_unlearning} represents our evaluation results on classification unlearning tasks. Generally, class-level unlearning shows varying robustness against unlearning mapping attacks, especially the strong performance of retraining methods. This not only provides empirical evidence that a model can satisfy the robust unlearning criteria but also validates that class-wise discriminative unlearning effectively aligns the unlearned model with a retrained model. However, we notice that as the dataset becomes more complex, from CIFAR10 to Tiny-ImageNet, both unlearning efficacy and robustness drop vastly. While unlearning's golden standard, the retraining method, is still capable of producing a reliable unlearning outcome, other unlearning methods struggle to keep up. Even though some maintain good unlearning efficacy, their robustness drops vastly.

Note that we also include Test Accuracy(TA) in our evaluation metrics. While TA does not indicate any robustness of an unlearning method, it does reflect the strength of the unlearning. Since most unlearning methods have some hyperparameters that can adjust their strength, with stronger unlearning resulting in lower test accuracy, one can perform very strong unlearning to have good unlearning performance (low UA and MIA) and very low test accuracy, which is impractical. Therefore, TA serves as a balancing metric, where lower TA can imply stronger unlearning, but the trade-off needs to be considered for robustness.

\begin{table*}[t]
    \centering
    \small
    \setlength{\tabcolsep}{3.5pt}
    \begin{tabular}{ccccccccccccc}
        \toprule
        \multirow{2}[2]{*}{}
        & \multicolumn{6}{c}{IS$\uparrow$}& \multicolumn{6}{c}{FID$\downarrow$}\\
        & \multicolumn{2}{c}{No Atk}& \multicolumn{2}{c}{8/255}& \multicolumn{2}{c}{Unbound}& \multicolumn{2}{c}{No Atk}& \multicolumn{2}{c}{8/255}& \multicolumn{2}{c}{Unbound}\\
        \cmidrule(lr){2-7} \cmidrule(lr){8-13}
        & R& F& R& F& R& F& R& F& R& F& R& F\\
        \hline\toprule
        Original& 6.21& 6.39& -& -& -& -& 96.12& 103.38& -& -& -& -\\
        I2I& 6.18& 2.79& 6.21& 6.22& 6.20& 6.35& 100.15& 306.43& 94.76& 114.16& 96.98& 110.29\\
        SalUn& 6.05& 2.42& 6.02& 6.11& 6.13& 6.27& 130.45& 330.79& 102.75& 133.82& 94.15& 108.44\\
        \bottomrule
    \end{tabular}
    \caption{IS and FID results for image-to-image generation unlearning. \textit{R} and \textit{F} stand for retain set and forget set. The attack strength is set to 0, 8/255, and unbound (where the noise strength is unlimited). Note that a higher IS score indicates better image quality, while a lower FID score reflects improved image fidelity.}
    \label{tab:image_generation}
\end{table*}

\begin{table*}[t]
    \centering
     \small
    \setlength{\tabcolsep}{8pt}
    \begin{tabular}{cccccc}
        \toprule
        & Church& Garbage Truck& Nudity& Parachute& Tench\\
        \hline\toprule
        \multirow{2}{*}{AdvUnlearn}& 0/0.08& 0/0.06& 0.076/0.356& 0.02/0.14& 0/0.06\\
                                & 93.94/100& 95.82/100& 72.91/100& 89.18/100& 95.54/100\\
        \hline
        \multirow{2}{*}{EraseDiff}& 0.06/0.76& 0.06/0.52& 0/0.059& 0.04/0.82& 0/0.12\\
                                & 45.72/38& 64.68/88.5& 95.37/100& 29.44/8.5& 92.94/100\\
        \hline
        \multirow{2}{*}{ESD}& 0.14/0.76& 0.02/0.44& 0.212/0.907& 0.06/0.8& 0/0.46\\
                                & 38.58/20.5& 73.98/100& 25.86/11& 44.46/33& 67.04/100\\
        \hline
        \multirow{2}{*}{FMN}& 0.52/0.96& 0.46/0.98& 0.881/1& 0.52/1& 0.36/1\\
                                & 6.78/0& 3.96/1& 0.39/0& 2.1/0& 6.42/2.5\\
        \hline
        \multirow{2}{*}{SalUn}& 0.1/0.68& 0.02/0.5& 0.017/0.246& 0.08/0.88& 0/0.24\\
                                & 47.16/33& 67.72/93& 84.71/100& 30.32/14& 88.4/100\\
        \hline
        \multirow{2}{*}{Scissorhands}& 0/0.08& 0/0.04& 0/0.119& 0.02/0.44& 0/0.18\\
                                & 96.36/100& 99.7/100& 92.35/100& 78.38/100& 91.58/100\\
        \hline
        \multirow{2}{*}{SPM}& 0.44/0.96& 0.04/0.82& 0.559/1& 0.26/0.96& 0.06/0.94\\
                                & 7.52/1& 37.36/23& 1.69/0& 12.34/3& 17.92/8.5\\
        \bottomrule
    \end{tabular}
    \caption{Empirical experiment on Text-to-Image unlearning methods. Five objects or concepts are selected to test unlearning robustness. The results recorded in each cell are \textbf{pre/post ASR} (first row) and \textbf{average/median attack steps} (second row).}
    \label{tab:t2i_unlearn}
\end{table*}


\textbf{I2I unlearning.} Table \ref{tab:image_generation} and Figure \ref{fig:mae_result1}, \ref{fig:mae_result2} present qualitative and quantitative results for image generation tasks. Even when the unlearned model avoids reconstructing images from the forget set, it can still generate well-restored images when given an adversarial input. Note that the attack can be unbounded for generation tasks since we only care about the model's output. However, as shown in both figures, even small noise is sufficient for a successful attack, exposing the model’s vulnerability. Moreover, the unlearning mapping attack does not significantly impact samples from the retain set, indicating that distinguishing between the forget and retain sets in the verification process is unnecessary.

\textbf{T2I unlearning.} We represent the results in Table \ref{tab:t2i_unlearn}. While ASR requires consideration of both pre-attack and post-attack performance, the average attack step provides a clearer and more distinct separation between methods. These results indicate that although many unlearning approaches demonstrate strong efficacy, their robustness against adversarial attacks remains limited. It is also worth noting that for both Image-to-Image and Text-to-Image unlearning, the retraining baseline is absent, as retraining is typically impractical due to the scale of the training data.

\textbf{Overall}, our results across discriminative, Image-to-Image, and Text-to-Image unlearnings reveal a consistent gap: while many unlearning methods achieve reasonable efficacy, they pay little attention to robustness against adversarial attacks. Consequently, existing approaches exhibit fundamental weaknesses once probed with adversarial strategies. This observation directly supports the motivation outlined in Section 3.2 and Definition 4, highlighting the necessity of Robust Unlearning as a core requirement rather than an optional property. We hope that future research will explicitly incorporate robustness considerations into the design of unlearning algorithms, moving toward solutions that are not only effective but also resilient against adversarial recovery.

\section{Conclusion}
We present RUB, the first comprehensive benchmark designed to evaluate unlearning algorithms under adversarial recovery attacks. RUB spans three vision tasks and supports standardized evaluation protocols, enabling a systematic comparison of unlearning robustness. Our proposed UMA formulation and its task-specific implementations allow empirical probing of residual knowledge in unlearned models. Extensive experiments across 15 representative methods reveal that current unlearning approaches often fail to fully erase sensitive information. We hope RUB provides a rigorous foundation for future research into robust, attack-resilient machine unlearning.


{
    \bibliographystyle{sty_file/ieeenat_fullname}
    \bibliography{reference}
}
\newpage
\appendix
\twocolumn[{
\centering
{\Large\bfseries RUB: Evaluating Residual Knowledge in Unlearned Models\par}
\vspace{0.5em}
{\large Supplementary Material\par}
\vspace{1em}
}]

\section{UMA implementation and ablations}
\subsection{Gradient-based implementation of UMA}
To achieve the attack goal outlined in (\ref{eq_goal}), we introduce a gradient-based input mapping attack: 
\begin{equation} \label{eq_opt}
    \mathop{\arg\min}_{\{\delta_x\}} E_{x\in\mathcal{D}_u} \left[D
    (f_u(\delta_x;\theta^u),f(x;\theta))\right],
\end{equation}
where $D$ quantifies the difference and can vary depending on the context, such as Mean Square Error Loss, Binary Cross-Entropy Loss, or KL Divergence Loss. To solve this optimization problem, we adopt the Projected Gradient Descent (PGD) method in \cite{madry2017towards} to find the input $\delta$. While PGD is normally used to maximize empirical loss, in this case, we aim to \textit{minimize} the loss, thus taking the opposite direction of the gradient update:
\begin{equation}
    \delta^{t+1}_x=\delta^t_x-\alpha\cdot sign[\nabla_{\delta_x} D(f_u(\delta^t_x;\theta^u),f(x;\theta)],
\end{equation}
where $\alpha$ stands for the step size for each iteration. The pseudocode of UMA is provided in Algorithm \ref{alg_general}. For simplicity, we only adopt the PGD-based mapping method as our baseline, though other optimization techniques can be substituted for potentially better performance.

{\color{black}{Though UMA and Robust Unlearning are consistent in principle, the optimization problem is typically non-convex. As a result, Algorithm~\ref{alg_general} does not guarantee exploration of all possible perturbations. Yet, it still provides a practical and actionable framework for identifying vulnerabilities in unlearning methods. Even in cases where UMA does not succeed, the absence of successful attacks strengthens the empirical evidence that the model may satisfy the robust unlearning criteria.}}

\begin{algorithm}[htbp]
\caption{Unlearning Mapping Attack}
\label{alg_general}
    \begin{algorithmic}[1]
    \small
        \STATE {\bfseries Input:} Pre-trained model $f(\cdot;\theta)$, Unlearned model $f_u(\cdot;\theta^u)$, 
        Unlearning dataset $\mathcal{D}_u$, Attack steps $T$, Attack step size $\eta$
        \STATE {\bfseries Output:} 
        Attack dataset $\mathcal{D}_{atk}$

        \STATE{Random initialize attack noise ${\{\delta_x\}}$ for $x\in \mathcal{D}_u$}
        \FOR{\texttt{$k$ = 0 \textbf{to} $T$}}
            \STATE{Calculate loss $\psi \gets \sum_{x\in\mathcal{D}_u}D(f(x;\theta), f_u(\delta_x^k;\theta^u))$}
            \STATE{Update attack noise ${\{\delta_x^{k+1}\}} \gets {\{\delta_x^k\}} - \eta \cdot sign(\nabla_{\delta_x}\psi)$}
            \STATE{$\{\delta_x^{k+1}\} \gets clip({\{\delta_x^{k+1}\}},0,1)$}
        \ENDFOR
        \STATE{Construct attack dataset $\mathcal{D}_{atk} \gets (\delta_x^{k+1}, y_x)$}
    \end{algorithmic}
\end{algorithm}

\subsection{Ablation Study}
We conduct ablation experiments on the two hyperparameters in UMA, the number of steps and step size. All experiments are done using discriminative models on CIFAR10 dataset. SalUn~\citep{fan2024salun} is chosen as the unlearning algorithm. All ablation experiments on step sizes have a fixed number of steps of 100, and all ablations on iteration numbers have a fixed step size of 1/255. Attack strength is set to 16/255 across all ablations.

As shown in Figure \ref{fig:ablation2}, the attack efficacy generally increases as the number of steps goes up. However, higher iteration numbers result in greater computation costs, which form a trade-off that the attacker needs to make. On the other hand, as shown in Figure \ref{fig:ablation1}, the attack step size reaches its best performance, around 0.7/255 to 1/255. A larger step size will cause the attack to find an incorrect direction, reducing the attack efficacy, while a smaller step size will generally cause a slow convergence speed, requiring a larger iteration step to reach equivalent performance. 


\begin{figure}[htbp]
    \centering
    \includegraphics[width=\linewidth]{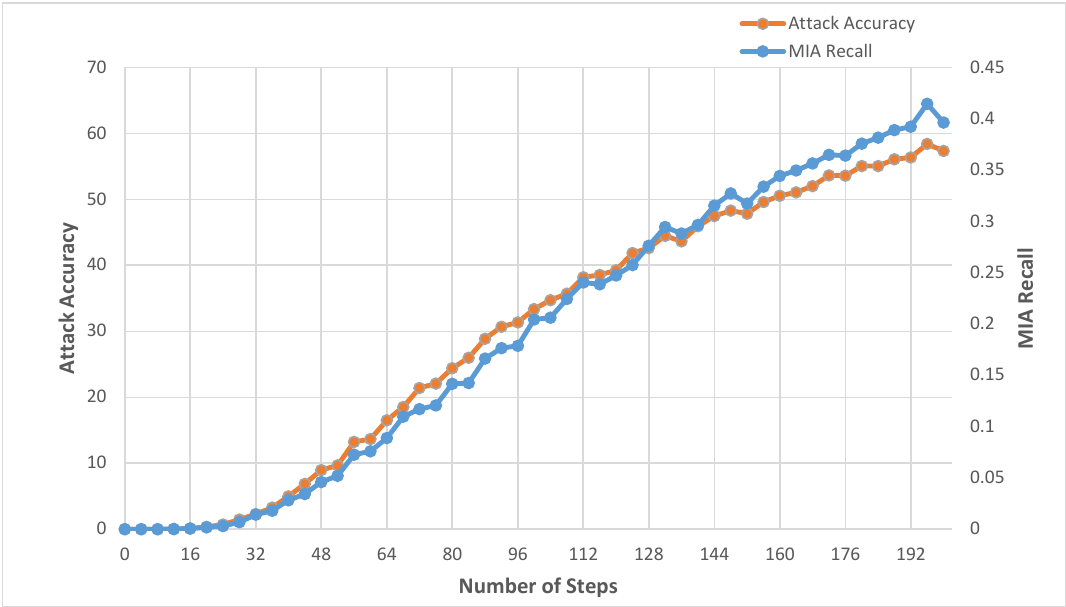}
    \caption{Ablation on attack iteration numbers. The experiments are done on CIFAR10 using SalUn~\citep{fan2024salun} as the baseline unlearning algorithm. All experiments have a fixed step size of 1/255 and an attack strength of 16/255.}
    \label{fig:ablation2}
\end{figure}

\begin{figure}[htbp]
    \centering
    \includegraphics[width=\linewidth]{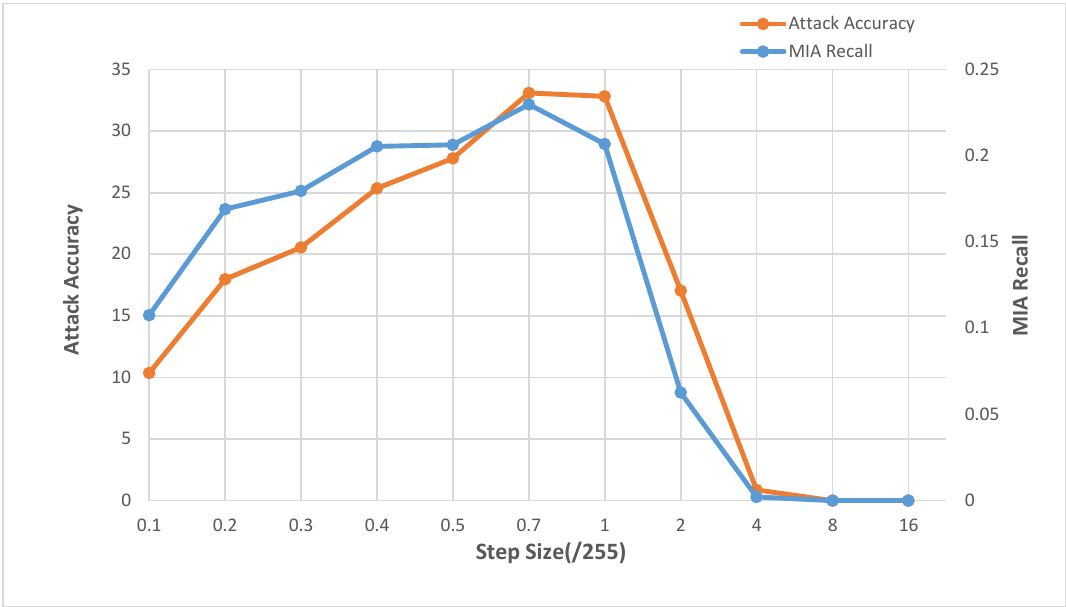}
    \caption{Ablation on attack step size. The experiments are done on CIFAR10 using SalUn~\citep{fan2024salun} as the baseline unlearning algorithm. All experiments have a fixed number of steps of 100 and an attack strength of 16/255.}
    \label{fig:ablation1}
\end{figure}

\section{Details on I2I unlearning setup}
In the experiments on image-to-image generative unlearning models, we evaluate whether our UMA attacks could explore the residue information left in the model after unlearning and resurface the "forgotten" knowledge. To this end, we follow the previous arts in I2I where the generative model is used to recover the masked region in a query image. To ease the discussion, let’s first clarify the data flow and pipeline of the generative model experiment. In our experiments, the generative unlearning pipeline involves the following steps:
\begin{itemize}
    \item $I_0$: The ground truth image from the forget set.
    \item $I_m$: The masked version of the image $I_0$, which serves as the input to the generative model.
    \item $I_1$: The output of the original generative model (before unlearning), where the masked regions in $I_m$ are reconstructed.
    \item $I_2$: The output of the unlearned generative model, which cannot reconstruct the masked regions for the forget set and instead generates gray or noisy outputs.
    \item $I_3$: The output of the unlearned generative model when attacked with UMA, which aims to resurface the forgotten information and reconstruct the masked regions as $I_1$.
\end{itemize}
By design, $I_1$, $I_2$, and $I_3$ are naturally different from the masked input $I_m$, as the goal of the generative model is to reconstruct the missing regions. Additionally, for the forget set, $I_2$ differs significantly from $I_1$, as the unlearned model is intended to "forget" the knowledge and cannot recover $I_0$ from $I_m$. UMA's goal is to probe whether the unlearned model can generate $I_3$ that closely resembles $I_1$, thereby bypassing the unlearning mechanism. Based on the above context, 
UMA’s efficacy is evaluated by how closely $I_3$ (the UMA output) resembles $I_1$ (the output of the original generative model before unlearning). This indicates whether the unlearned model retains residual knowledge of the forget set, effectively failing to fully "forget."

To verify UMA's impact, we directly computed the L1 distance between $I_3$  and $I_1$ per image. As shown in the Table~\ref{tab:L1Norm}, the L1 differences between $I_1$ and $I_3$ are very small after the attack (e.g. for the 224x224x3 image, average 0.3 intensity difference per pixel for the forget set with I2I~\citep{li2024machine} and 1.6 intensity difference per pixel for the SalUn~\citep{fan2024salun}), indicating that UMA can prompt the unlearned model to output information it was supposed to forget. This provides strong evidence that UMA effectively bypasses the unlearning process.

In addition, we include multiple visual examples in Figure~\ref{fig:mae_extend1} and ~\ref{fig:mae_extend2}. These examples present images for $I_0$, $I_m$, $I_1$, $I_2$, and $I_3$, providing a clear comparison of the reconstruction results across all stages of the pipeline. These visualizations demonstrate how UMA successfully recovers information that should have been forgotten, illustrating its effectiveness in attacking the unlearning mechanism.

\begin{table}
    \centering
    \small
    \begin{tabular}{ccccc}
        \toprule
        \multirow{2}[2]{*}{L1 per image}
        & \multicolumn{2}{c}{ISI~\citep{li2024machine}}& \multicolumn{2}{c}{SalUn~\citep{fan2024salun}}\\
        \cmidrule(lr){2-3} \cmidrule(lr){4-5}
        & No Attack & 8/255  & No Attack& 8/255 \\
        \hline\toprule
        Retain set& 64,619& 42,410& 214,596& 114,089\\
        Forget set& 1,140,778& 48,317& 2,790,552& 242,029\\
        \bottomrule
    \end{tabular}
    \caption{L1 norm between the outputs of the generative model before and after unlearning. The values under no attack are calculated by $L1(I_2, I_1)$, and the values under the attack strength 8/255 are computed by $L1(I_3,I_1)$. }
    \label{tab:L1Norm}
\end{table}

\begin{figure*}[htbp]
    \centering
    \includegraphics[width=\linewidth]{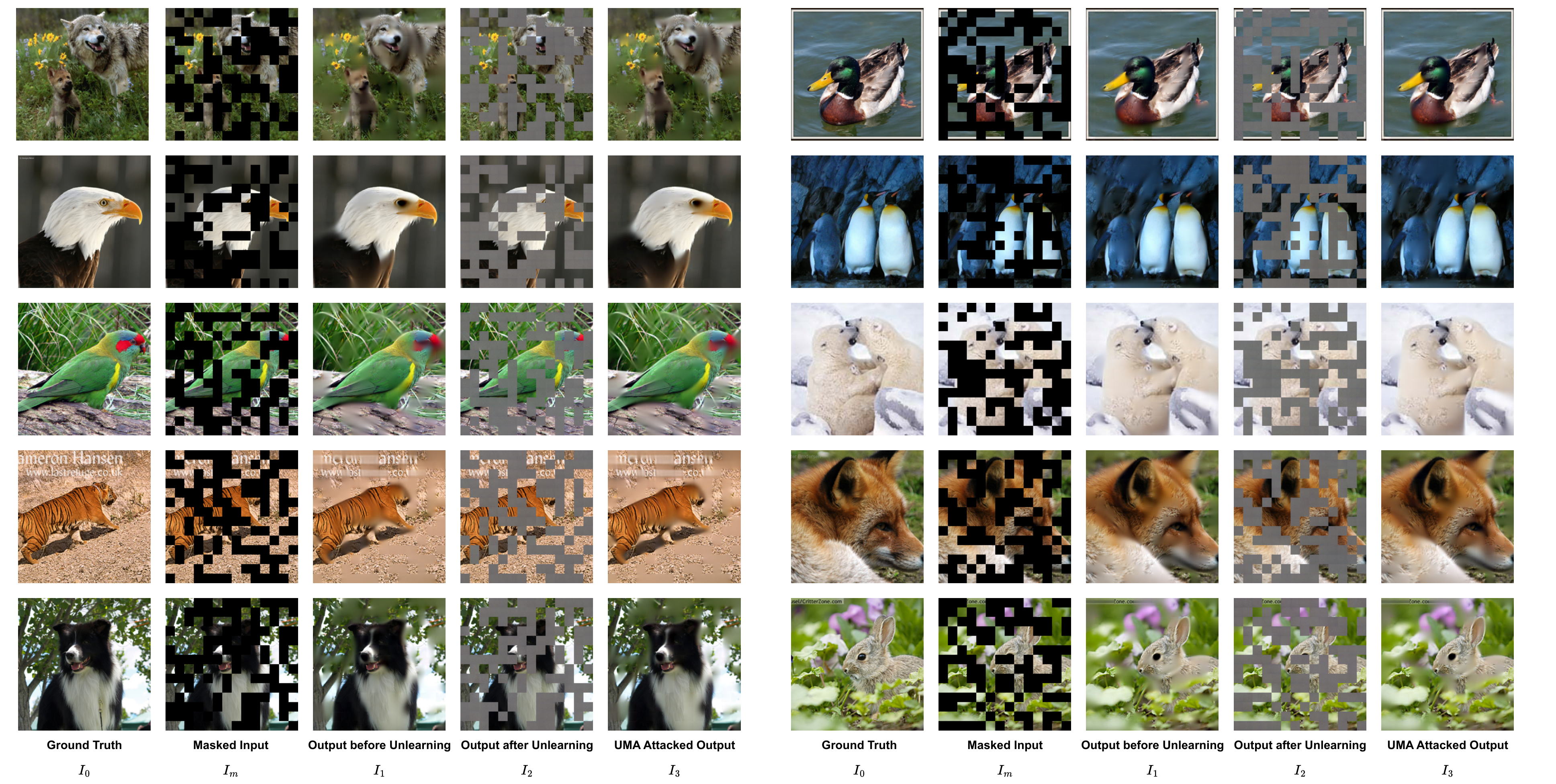}
    \caption{Examples of the generated images using I2I~\citep{li2024machine} unlearning methods. Ground truth, $I_0$, Masked Input, $I_m$, Output before Unlearning, $I_1$, Output after Unlearning, $I_2$, UMA Attacked Output, $I_3$, are represented here as discussed in Section A.3}
    \label{fig:mae_extend1}
\end{figure*}
\begin{figure*}[htbp]
    \centering
    \includegraphics[width=\linewidth]{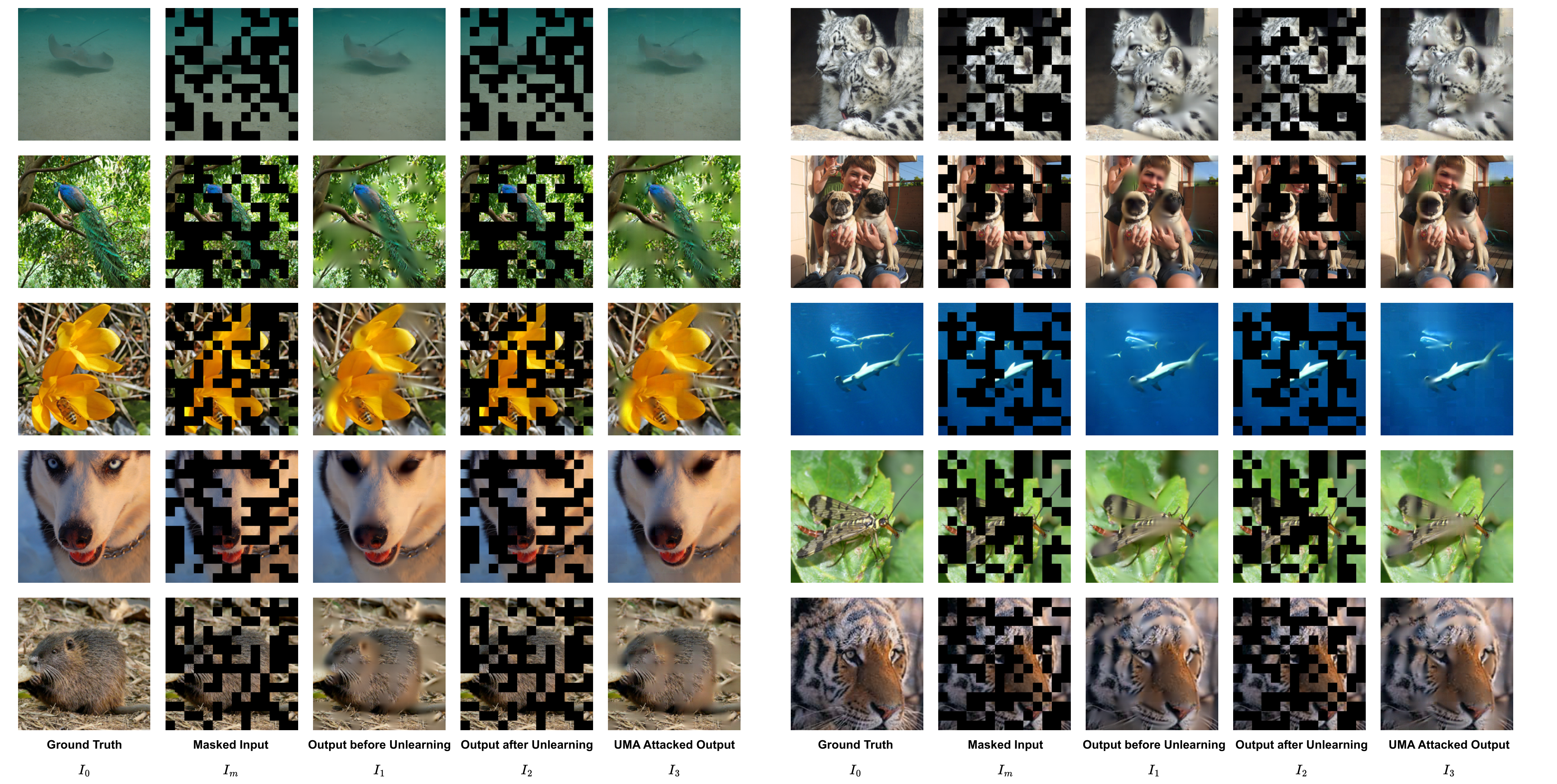}
    \caption{Examples of the generated images using SalUn~\citep{fan2024salun} unlearning methods. Ground truth, $I_0$, Masked Input, $I_m$, Output before Unlearning, $I_1$, Output after Unlearning, $I_2$, UMA Attacked Output, $I_3$, are represented here as discussed in Section A.3}
    \label{fig:mae_extend2}
\end{figure*}

\section{Experimental Evaluation on discriminative unlearning}

\subsection{MIA Implementation}
For the MIA evaluation of discriminative unlearning, we adopt a shadow-model-based MIA strategy~\cite{shokri2017membership} for the quantitative measurement. Specifically, 10\% of the total dataset is randomly sampled to train 10 shadow models, each implemented as a ResNet50 and trained for 10 epochs(20 epochs for Tiny-ImageNet). We then collect the logit outputs of these shadow models on both their seen and unseen data to construct the shadow dataset. Using this dataset, we train simple attack models designed to determine whether a given logit is from seen or unseen data. To ensure fine-grained discrimination, we employ one independent attack model per class. Finally, the attack recall is recorded and reported. To address the randomness in MIA for reliable evaluation, we randomly sampled 10 fixed random seeds for executing unlearning and 5 fixed random seeds for training MIA attack models. In total, we have 50 sets of results, and their average and standard deviation are reported in Table \ref{tab:class_unlearning_full}.

\begin{table*}[t]
    \centering
    \small
    \setlength{\tabcolsep}{2pt}
    \begin{tabular}{cccccccc}
        \toprule
        \multirow{2}[2]{*}{\textbf{CIFAR10}}
        & \multicolumn{3}{c}{No Atk}& \multicolumn{2}{c}{$\epsilon=8/255$}& \multicolumn{2}{c}{$\epsilon=16/255$}\\
        \cmidrule(lr){2-4} \cmidrule(lr){5-6} \cmidrule(lr){7-8}
        & TA$\uparrow$& UA$\downarrow$& MIA$\downarrow$& UA$\downarrow$& MIA$\downarrow$& UA$\downarrow$& MIA$\downarrow$\\
        \hline\toprule
        Original& 94.13& 100& 0.9796& -& -& -& -\\
        retrain& $94.14_{\pm0.20}$& $0_{\pm0}$& $0_{\pm0}$& $0_{\pm0}$& $0_{\pm0}$& $0_{\pm0}$& $0_{\pm0}$\\
        FT& $91.82_{\pm0.42}$& $20.55_{\pm4.18}$& $0.088_{\pm0.029}$& $99.96_{\pm0.03}$& $0.995_{\pm0.034}$& $99.98_{\pm0.02}$& $0.995_{\pm0.003}$\\
        RL& $92.19_{\pm0.43}$& $0_{\pm0}$& $0_{\pm0}$& $5.82_{\pm5.23}$& $0.024_{\pm0.024}$& $26.64_{\pm18.28}$& $0.135_{\pm0.111}$\\
        IU& $88.06_{\pm2.51}$& $8.76_{\pm5.70}$& $0.058_{\pm0.040}$& $99.12_{\pm0.38}$& $0.965_{\pm0.012}$& $99.87_{\pm0.09}$& $0.983_{\pm0.008}$\\
        $l_1$-sparse& $90.00_{\pm0.16}$& $0_{\pm0.01}$& $0_{\pm0}$& $98.30_{\pm0.41}$& $0.871_{\pm0.081}$& $99.90_{\pm0.04}$& $0.978_{\pm0.018}$\\
        SalUn& $92.70_{\pm0.25}$& $0_{\pm0}$& $0_{\pm0}$& $7.13_{\pm9.59}$& $0.036_{\pm0.062}$& $26.87_{\pm16.39}$& $0.148_{\pm0.150}$\\
        \toprule
        \multirow{2}[2]{*}{\textbf{CIFAR100}}
        & \multicolumn{3}{c}{No Atk}& \multicolumn{2}{c}{$\epsilon=8/255$}& \multicolumn{2}{c}{$\epsilon=16/255$}\\
        \cmidrule(lr){2-4} \cmidrule(lr){5-6} \cmidrule(lr){7-8}
        & TA$\uparrow$& UA$\downarrow$& MIA$\downarrow$& UA$\downarrow$& MIA$\downarrow$& UA$\downarrow$& MIA$\downarrow$\\
        \hline\toprule
        Original& 75.25& 100& 0.9908& -& -& -& -\\
        retrain& $75.40_{\pm1.04}$& $0_{\pm0}$& $0.024_{\pm0.015}$& $0_{\pm0}$& $0.014_{\pm0.010}$& $0_{\pm0}$& $0.014_{\pm0.011}$\\
        FT& $67.64_{\pm1.22}$& $0.48_{\pm0.26}$& $0.306_{\pm0.060}$& $99.28_{\pm0.17}$& $0.977_{\pm0.031}$& $99.89_{\pm0.03}$& $0.992_{\pm0.015}$\\
        RL& $69.96_{\pm0.51}$& $3.20_{\pm2.88}$& $0.269_{\pm0.060}$& $51.53_{\pm3.62}$& $0.688_{\pm0.080}$& $80.50_{\pm3.08}$& $0.790_{\pm0.074}$\\
        IU& $66.42_{\pm2.47}$& $53.37_{\pm7.11}$& $0.848_{\pm0.054}$& $99.93_{\pm0.03}$& $1_{\pm0}$& $99.93_{\pm0.02}$& $1_{\pm0}$\\
        $l_1$-sparse& $70.70_{\pm0.61}$& $1.30_{\pm0.25}$& $0.402_{\pm0.099}$& $99.77_{\pm0.05}$& $0.925_{\pm0.056}$& $99.91_{\pm0.03}$& $0.945_{\pm0.049}$\\
        SalUn& $73.89_{\pm0.34}$& $4.13_{\pm3.55}$& $0.221_{\pm0.038}$& $61.57_{\pm5.04}$& $0.788_{\pm0.045}$& $85.16_{\pm3.43}$& $0.888_{\pm0.035}$\\
        \toprule
        \multirow{2}[2]{*}{\textbf{Tiny-ImageNet}}
        & \multicolumn{3}{c}{No Atk}& \multicolumn{2}{c}{$\epsilon=8/255$}& \multicolumn{2}{c}{$\epsilon=16/255$}\\
        \cmidrule(lr){2-4} \cmidrule(lr){5-6} \cmidrule(lr){7-8}
        & TA$\uparrow$& UA$\downarrow$& MIA$\downarrow$& UA$\downarrow$& MIA$\downarrow$& UA$\downarrow$& MIA$\downarrow$\\
        \hline\toprule
        Original& 64.17& 99.96& 1& -& -& -& -\\
        retrain& $57.74_{\pm0.67}$& $0_{\pm0}$& $0_{\pm0}$& $0_{\pm0}$& $0_{\pm0}$& $0_{\pm0}$& $0_{\pm0}$\\
        FT& $60.48_{\pm0.19}$& $79.01_{\pm0.69}$& $0.721_{\pm0.014}$& $99.99_{\pm0.01}$& $0.991_{\pm0.004}$& $99.99_{\pm0.01}$& $0.991_{\pm0.003}$\\
        RL& $56.23_{\pm0.31}$& $2.09_{\pm0.31}$& $0.028_{\pm0.009}$& $99.78_{\pm0.11}$& $0.859_{\pm0.023}$& $99.99_{\pm0.01}$& $0.917_{\pm0.028}$\\
        IU& $57.71_{\pm1.82}$& $94.44_{\pm4.26}$& $0.882_{\pm0.052}$& $99.99_{\pm0.01}$& $0.991_{\pm0.004}$& $99.99_{\pm0.01}$& $0.991_{\pm0.003}$\\
        $l_1$-sparse& $58.28_{\pm0.35}$& $45.99_{\pm0.67}$& $0.228_{\pm0.028}$& $99.99_{\pm0.01}$& $0.833_{\pm0.017}$& $99.99_{\pm0.01}$& $0.843_{\pm0.019}$\\
        SalUn& $57.82_{\pm0.15}$& $5.95_{\pm0.78}$& $0.084_{\pm0.019}$& $99.98_{\pm0.01}$& $0.964_{\pm0.017}$& $99.99_{\pm0.01}$& $0.982_{\pm0.010}$\\
        \bottomrule
    \end{tabular}
    \caption{Full evaluation of Test Accuracy (TA), Unlearning Accuracy (UA), and MIA scores before and after Unlearning Mapping Attack for the Class Unlearning scenario. Attack is bounded with 8/255 and 16/255. The original here indicates the model performance before unlearning.}
    \label{tab:class_unlearning_full}
\end{table*}

\subsection{Instance-level unlearning Results}
For instance-level classification unlearning, as shown in Table \ref{tab:instance_unlearning}, all baseline methods display limited robustness against unlearning mapping attacks. While the retraining method performs the best, it still lacks sufficient robustness, even with $\epsilon=8/255$. This suggests that attackers can easily manipulate unlearned images, causing the model to re-recognize them, thus compromising the unlearning process.
\begin{table*}[t]
    \centering
    \small
    \setlength{\tabcolsep}{10pt}
    \begin{tabular}{cccccccc}
        \toprule
        \multirow{2}[2]{*}{\textbf{CIFAR10}}
        & \multicolumn{3}{c}{No Atk}& \multicolumn{2}{c}{8/255}& \multicolumn{2}{c}{16/255}\\
        \cmidrule(lr){2-4} \cmidrule(lr){5-6} \cmidrule(lr){7-8}
        & TA& UA& MIA& UA& MIA& UA& MIA\\
        \hline\toprule
        Original& 94.13& 100& 0.9732& -& -& -& -\\
        retrain& 93.34& 93.78& 0.8636& 99.98& 0.9774& 99.98& 0.9728\\
        FT& 92.13& 98.02& 0.9124& 99.98& 0.9794& 99.96& 0.9810\\
        RL& 89.22& 91.88& 0.8012& 99.96& 0.9896& 100& 0.9866\\
        IU& 89.82& 97.92& 0.8926& 99.98& 0.9630& 99.98& 0.9628\\
        $l_1$-sparse& 91.32& 95.76& 0.8848& 99.98& 0.9842& 100& 0.9814\\
        SalUn& 90.55& 93.48& 0.8140& 100& 0.9884& 99.98& 0.9872\\
        \toprule
        \multirow{2}[2]{*}{\textbf{CIFAR100}}
        & \multicolumn{3}{c}{No Atk}& \multicolumn{2}{c}{8/255}& \multicolumn{2}{c}{16/255}\\
        \cmidrule(lr){2-4} \cmidrule(lr){5-6} \cmidrule(lr){7-8}
        & TA& UA& MIA& UA& MIA& UA& MIA\\
        \hline\toprule
        Original& 75.25& 100& 0.9924& -& -& -& -\\
        retrain& 73.92& 72.72& 0.7354& 99.92& 0.9910& 100& 0.9932\\
        FT& 70.90& 96.44& 0.9436& 99.96& 0.9970& 100& 0.9972\\
        RL& 71.05& 86.04& 0.7786& 99.98& 0.9946& 100& 0.9956\\
        IU& 71.89& 99.20& 0.9702& 100& 0.9894& 100& 0.9912\\
        $l_1$-sparse& 69.60& 90.10& 0.7404& 99.98& 0.9704& 99.98& 0.9756\\
        SalUn& 71.99& 88.72& 0.7936& 99.94& 0.9890& 100& 0.9914\\
        \toprule
        \multirow{2}[2]{*}{\textbf{Tiny-ImageNet}}
        & \multicolumn{3}{c}{No Atk}& \multicolumn{2}{c}{8/255}& \multicolumn{2}{c}{16/255}\\
        \cmidrule(lr){2-4} \cmidrule(lr){5-6} \cmidrule(lr){7-8}
        & TA& UA& MIA& UA& MIA& UA& MIA\\
        \hline\toprule
        Original& 64.17& 99.98& 0.9978& -& -& -& -\\
        retrain& 61.81& 60.17& 0.6387& 99.97& 0.9735& 100& 0.9811\\
        FT& 55.66& 85.42& 0.8908& 99.99& 0.9969& 99.97& 0.9969\\
        RL& 55.36& 72.88& 0.8002& 99.99& 0.9962& 99.98& 0.9968\\
        IU& 56.33& 94.85& 0.9591& 99.97& 0.9967& 99.98& 0.9969\\
        $l_1$-sparse& 56.04& 61.71& 0.3836& 99.99& 0.7597& 100& 0.7614\\
        SalUn& 54.94& 66.99& 0.6237& 99.99& 0.9654& 99.99& 0.9681\\
        \bottomrule
    \end{tabular}
    \caption{Test Accuracy (TA), Unlearning Accuracy (UA), and MIA scores before and after Unlearning Mapping Attack for the Instance Unlearning scenario. Attack is bounded with 8/255 and 16/255. The original here indicates the model performance before unlearning.}
    \label{tab:instance_unlearning}
\end{table*}

\section{Hyperparameter Settings of our Pre-Trained Unlearnings}
We perform the \textbf{Discriminative Unlearning} and \textbf{Image-to-Image Unlearning} by ourselves. Specifically, we adopt the discriminative unlearning code from SalUn's project, https://github.com/OPTML-Group/Unlearn-Saliency, and we use code from I2I~\citep{li2024machine}, https://github.com/jpmorganchase/i2i\_mage/tree/i2i, as reference when constructing Image-to-Image unlearning evaluations, as well as the unlearning class index when performing Image-to-Image Unlearning. We list our detailed hyperparameter selection for discriminative unlearning in Table \ref{tab:unlearn_hyperparameter}. For Image-to-Image Unlearning, we use the SGD optimizer, learning rate 0.01 for Salun, and the AdamW optimizer, base learning rate 1e-4 for I2I. For the \textbf{Text-to-Image} task, we utilize the unlearned model checkpoint from UnlearnDiffAtk~\citep{zhang2024generate}'s project page, https://github.com/OPTML-Group/Diffusion-MU-Attack, for evaluation.
\begin{table}[t]
    \centering
    \small
    \setlength{\tabcolsep}{3.5pt}
    \begin{tabular}{cccc}
        \toprule
        & CIFAR10& CIFAR100& Tiny-ImageNet\\
        \hline\toprule
        FT& \makecell[l]{epoch=10\\ lr=0.013}& \makecell[l]{epoch=10\\ lr=0.013}& \makecell[l]{epoch=10\\ lr=0.0023}\\
        \hline
        RL& \makecell[l]{epoch=10\\ lr=0.013}& \makecell[l]{epoch=10\\ lr=0.02}& \makecell[l]{epoch=10\\ lr=0.0025}\\
        \hline
        IU& $\alpha=20$& $\alpha=20$& $\alpha=10$\\
        \hline
        $l_1$-sparse& \makecell[l]{epoch=10\\ lr=0.001\\ $\alpha=0.001$}& \makecell[l]{epoch=10\\ lr=0.001\\ $\alpha=0.0007$}& \makecell[l]{epoch=10\\ lr=0.001\\ $\alpha=0.0001$}\\
        \hline
        SalUn& \makecell[l]{epoch=10\\ lr=0.013}& \makecell[l]{epoch=10\\ lr=0.022}& \makecell[l]{epoch=10\\ lr=0.0015}\\
        \bottomrule
    \end{tabular}
    \caption{Detailed hyperparameters used for discriminative unlearning evaluations.}
    \label{tab:unlearn_hyperparameter}
\end{table}

\end{document}